\documentclass{article} 
\usepackage{hyperref}
\usepackage{aaai24}  
\usepackage{times}  
\usepackage{helvet}  
\usepackage{courier}  
\usepackage{url}  
\usepackage{graphicx} 
\usepackage{natbib}  
\usepackage{caption} 
\usepackage{algorithm}
\usepackage{algorithmic}
\usepackage{adjustbox}
\usepackage{multirow}

\usepackage{newfloat}
\usepackage{listings}

\usepackage{amsmath}
\usepackage{amssymb}
\usepackage{xcolor}

\title{Securing Neural Networks with Knapsack Optimization}
\author {
    Yakir Gorski,
    Amir Jevnisek,
    Shai Avidan
}
\affiliations {
     yakirg320@gmail.com, amirjevn@mail.tau.ac.il, avidan@eng.tau.ac.il
}

\usepackage{bibentry}

\begin{document}

\maketitle

\begin{abstract}
MLaaS Service Providers (SPs) holding a Neural Network would like to keep the Neural Network weights secret.  On the other hand, users wish to utilize the SPs' Neural Network for inference without revealing their data. Multi-Party Computation (MPC) offers a solution to achieve this. Computations in MPC involves communication, as the parties send data back and forth. Non-linear operations are usually the main bottleneck requiring the bulk of communication bandwidth. In this paper, we focus on ResNets, which serve as the backbone for many Computer Vision tasks, and we aim to reduce their non-linear components, specifically, the number of ReLUs.

Our key insight is that spatially close pixels exhibit correlated ReLU responses. Building on this insight, we replace the per-pixel ReLU operation with a ReLU operation per patch. We term this approach 'Block-ReLU'. Since different layers in a Neural Network correspond to different feature hierarchies, it makes sense to allow patch-size flexibility for the various layers of the Neural Network. We devise an algorithm to choose the optimal set of patch sizes through a novel reduction of the problem to the Knapsack Problem. We demonstrate our approach in the semi-honest secure 3-party setting for four problems: Classifying ImageNet using ResNet50 backbone, classifying CIFAR100 using ResNet18 backbone, Semantic Segmentation of ADE20K using MobileNetV2 backbone, and Semantic Segmentation of Pascal VOC 2012 using ResNet50 backbone. Our approach achieves competitive performance compared to a handful of competitors. Our source code is publicly available: \url{https://github.com/yg320/secure_inference}.

\end{abstract}

\begin{figure}[t]
\begin{tabular}{c|c}
    
   CIFAR-100 & Pascal VOC 2012 \\
   \includegraphics[width=0.45\linewidth]{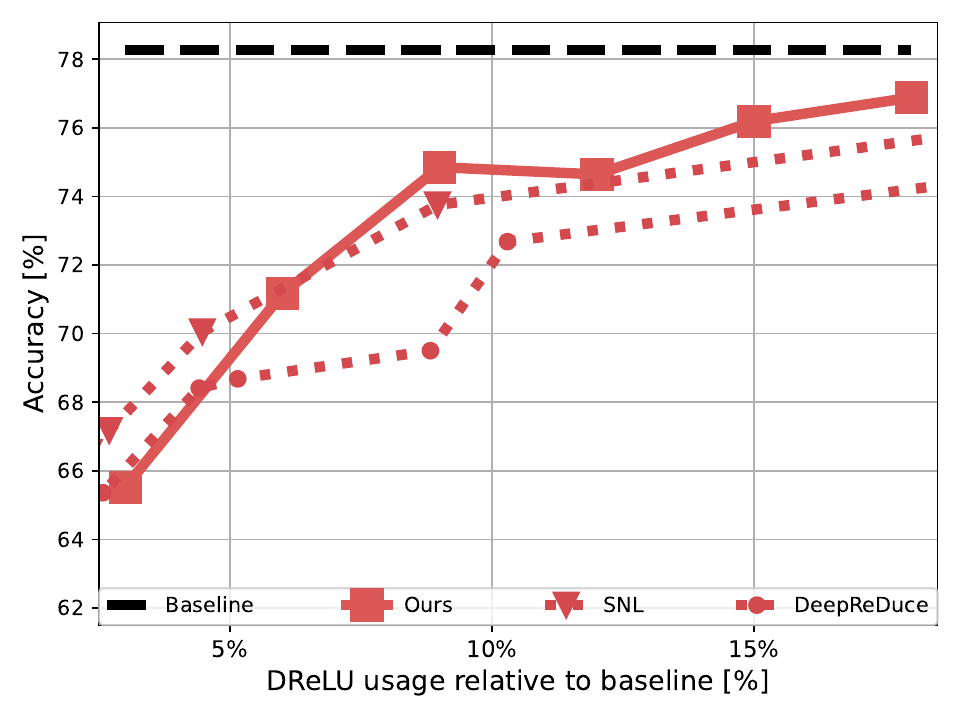}  & 
   \includegraphics[width=0.45\linewidth]{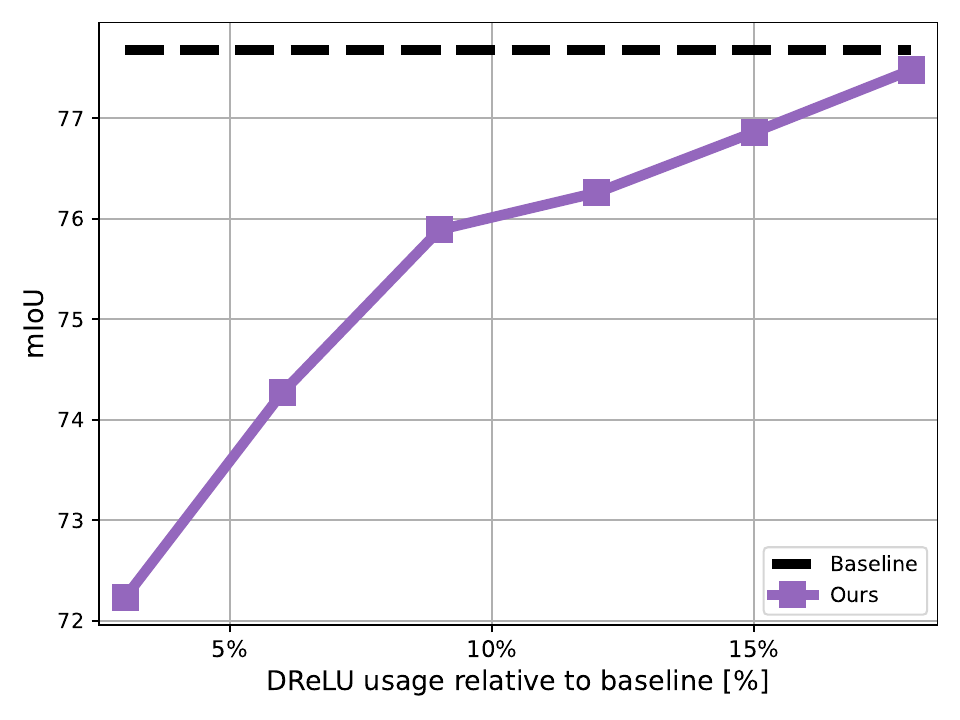}
   \\
   ImageNet & ADE20K \\
   \includegraphics[width=0.45\linewidth]{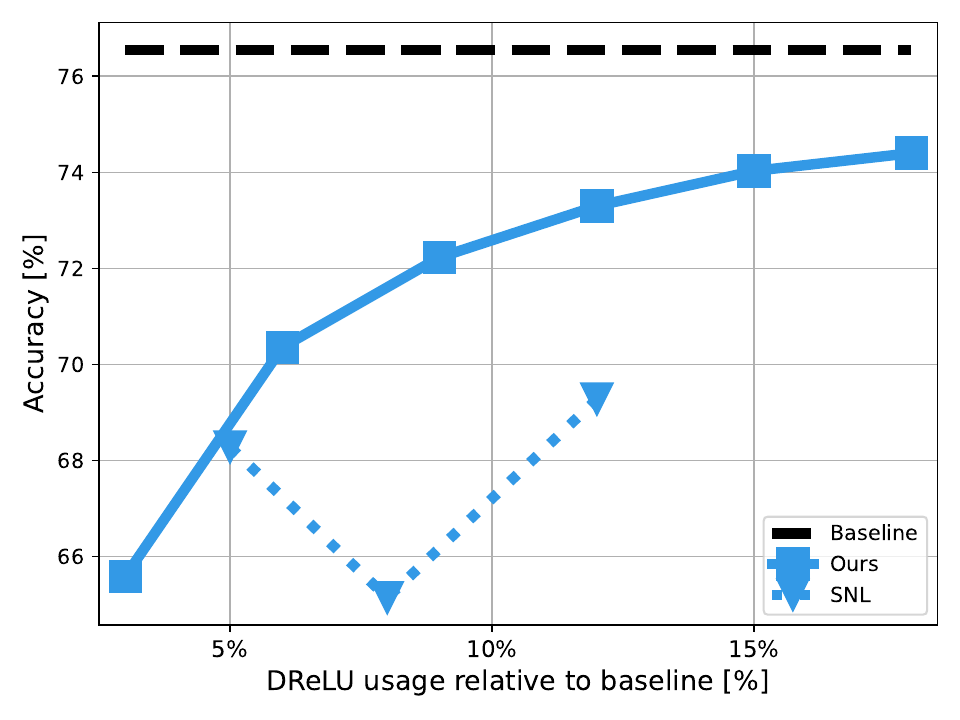}  & 
   \includegraphics[width=0.45\linewidth]{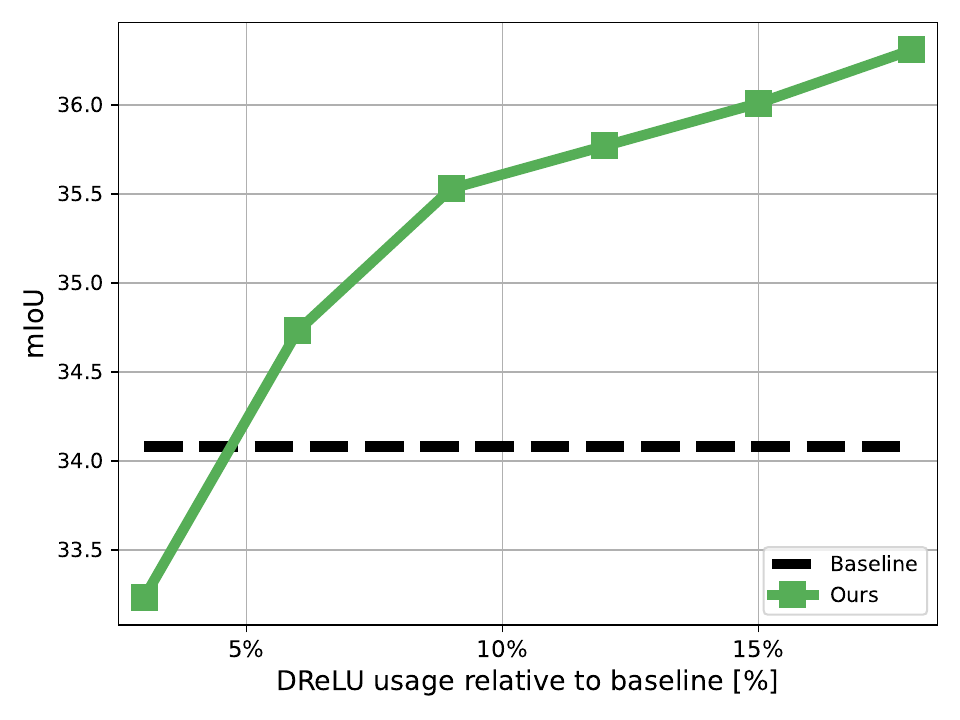}
   \\
   (a) Classification & (b) Semantic Segmentation \\
\end{tabular}
   \caption{{\bf Task Performance vs DReLU budget:} (a) Classification results for CIFAR-100 and ImageNet. We show competitive results with respect to the alternatives. (b) Semantic Segmentation results. There are no published results for Secure Semantic Segmentation to compare against.
   }
\label{fig:teaser}
\end{figure}

\section{Introduction}
With the recent surge in the popularity of machine learning algorithms, there is a similar increase in the interest of Private Inference (PI). In a typical PI scenario, two parties, a client with an image and an ML-provider server with a deep-learning model, seek to collaborate for the purpose of running inference. However, privacy concerns often impede such cooperation as both the client’s image and the server’s model may contain information that neither party is willing to share. 

Secure Inference protocols enable Private Inference, yet they come at the expense of increased runtime and communication bandwidth consumption due to the necessary communication rounds for inference evaluation. Non-linear functions particularly contribute to these communication rounds. Neural Networks involve both linear and non-linear operations. For example, ResNet18 classifying CIFAR-100 contains over $500 \mathrm{K}$ ReLU activations. ReLUs are the product of the input and a non-linear function which is termed DReLU. DReLU is an indicator function which is set to $1$  if the input is positive and is $0$ when the input is negative. The goal of this paper is to reduce the number of DReLUs in a Neural Network. 

The trade-off between reducing DReLUs in the network and task performance is illustrated in Figure~\ref{fig:teaser}. In Figure~\ref{fig:teaser}(a), we demonstrate competitive results against current state-of-the-art methods in the context of classifying CIFAR-100 with ResNet18. Furthermore, our approach outperforms the current SOTA method in classifying larger images like ImageNet. Additionally, we present the \emph{first} evaluation of secure inference for Semantic Segmentation, as depicted in Figure~\ref{fig:teaser}(b).

Secure inference aims for privacy-preserving computation in scenarios where a service provider and a client jointly perform calculations without exposing their data. While Homomorphic Encryption (HE) theoretically allows direct work on encrypted data, its real-world speed is impractical. Secure inference typically employs Multi-Party Computation (MPC), involving rounds of communication. However, a central challenge lies in the high computational cost of comparison operations like DReLU, which tends to dominate runtime and bandwidth.

Our main insight lies in the strong correlation of DReLU outcomes among neighboring pixels. To leverage this, we propose a strategy of employing a single ReLU operation per patch. We term ReLU per patch bReLU. However, a crucial question arises: what should be the optimal patch size? To address this we use a novel, Knapsack-based optimization strategy, to find an optimal configuration of patch sizes for all layers in the network.

Our main objective is to minimize DReLU counts while causing as little degradation to the network's performance as possible. Intuitively, we view this tradeoff as a variant of the Rate-Distortion tradeoff~\cite{berger2003rate}. We devise a network distortion measure, serving as a proxy for evaluating the network performance degradation under the replacement of ReLUs with bReLUs. Utilizing this proxy, we formulate our problem as an optimization goal with constraints. Given a DReLU count budget, our aim is to minimize the network's degradation while adhering to the budget. This problem aligns with the Multiple-Choice Knapsack Problem~\cite{kellerer2004multiple}, and we implement a pseudo-polynomial time Dynamic Programming solution that allows us to find the optimal set of patch sizes based on our assumptions.

Additionally, we implement bit truncation~\cite{nishide2007multiparty, ghodsi2021circa} to lower the cost of comparison operations. We achieve this by disregarding some of the most and least significant bits of the activation layers. To summarize, our method comprises four key steps: (1) quantifying the distortion of the Neural Network for each layer for a set of patch-size hypothesises, (2) solving for the Block-ReLU patch sizes using a Knapsack-based optimization strategy, (3) replacing ReLUs with Block-ReLUs to reduce non-linear operations in the network and fine-tune the network, and finally (4) employing bit truncation to further reduce computations and achieve a speedup.

Existing benchmarks for Secure Inference primarily focus on small images like CIFAR-100's $32 \times 32$ resolution or Tiny-ImageNet's $64 \times 64$, mostly limited to Classification tasks. To extend these benchmarks, we evaluate Private Inference on large images like ImageNet, which includes images with an average resolution of $469 \times 387$, and we are the \emph{first} to evaluate on Semantic Segmentation benchmarks like ADE20K and Pascal VOC12. We evaluate our model both on classical CIFAR-100 classification with ResNet18, as well as on ImageNet classification using ResNet50 and Semantic Segmentation of ADE20K using DeepLabV3 with MobileNetV2 backbone, and Semantic Segmentation of Pascal VOC 2012 using DeepLabV3 with ResNet50 backbone.

Since ReLU operations contribute the most to communication bandwidth, we save a considerable amount of network traffic. To validate this, we implement a semi-honest secure 3-party setting of the SecureNN protocol. We use $3$ instances within the same AWS EC2 region. ImageNet classification is executed in about $5.5$ seconds, and CIFAR100 classification in about $1.5$ seconds. ADE20K Semantic Segmentation is executed in about $32$ seconds and Pascal VOC 2012 on $85$ seconds. Classification and Segmentation performances measured are all comparable to OpenMMLab's non-secure models~\cite{mmseg2020, 2020mmclassification}.

To summarize, our main contributions are:

\begin{itemize}

\item We develop a generic, Knapsack-based, data-driven algorithm that greatly reduces the number non-linear operations (DReLUs) in the network. 

\item We demonstrate a significant runtime reduction with a marginal trade-off in accuracy across classical benchmarks for both low-resolution, as well as in the previously unaddressed large images, and the task of Semantic Segmentation.

\item We present a secure semantic segmentation algorithm that preserves the model accuracy while being $\times 10$ faster than a comparable secure baseline protocol.

\item We build and release an optimized, a purely Pythonic, wrapper code over OpenMMLab based packages that secures models taken from their model zoo.

\end{itemize}
\section{Related Work}

\paragraph{Privacy Preserving Deep Learning}
The research on privacy preserving deep learning shows significant differences across various aspects. These include the number of parties involved, threat models (Semi-honest, Malicious), supported layers, techniques used (e.g., Homomorphic encryption, garbled circuits, oblivious transfer, and secret sharing) and the capabilities provided (training and inference).

The pioneers in the field of performing prediction with neural networks on encrypted data were CryptoNets~\cite{gilad2016cryptonets}. They employed leveled homomorphic encryption, replacing ReLU non-linearities with square activations in order to perform inference on ciphertext. SecureML~\cite{mohassel2017secureml} utilized three distinct sharing methods: Additive, Boolean, and Yao sharing, along with a protocol to facilitate conversions between them. They used linearly homomorphic encryption (LHE) and oblivious transfer (OT) to precompute Beaver's triplets. MiniONN~\cite{liu2017oblivious} proposed to generate Beaver's triplets using additively homomorphic encryption together with the single instruction multiple data (SIMD) batch processing technique. GAZELLE~\cite{juvekar2018gazelle} suggested to use packed additively homomorphic encryption (PAHE) to make linear layers more communication efficient. FALCON~\cite{li2020falcon} achieved high efficiency by running convolution in the frequency domain, using Fast Fourier Transform (FFT) based ciphertext calculation. Chameleon~\cite{riazi2018chameleon} builds upon ABY~\cite{patra2021aby2} and employs a Semi-honest Third Party (STP) dealer to generate Beaver's triplets in an offline phase. SecureNN~\cite{wagh2019securenn} suggested three-party computation (3PC) protocols for secure evaluation of deep learning components. The Porthos component of CrypTFlow~\cite{kumar2020cryptflow} is an improved semi-honest 3-party MPC protocol that builds upon SecureNN. {\large F}ALCON ~\cite{wagh2020falcon}, combines techniques from SecureNN and ABY$^3$~\cite{mohassel2018aby3}, replacing the MSB and Share Convert protocols in SecureNN with cheaper Wrap$_3$ protocols.

\paragraph{ReLU Savings}
There are several methods to reduce ReLU expense through non-cryptographic means. One family of such methods, replaces non-linear functions with polynomial approximations for the non-linearities~\cite{hesamifard2018privacy, khan2021blind}. Another family of methods leverage Network Architecture Search (NAS) algorithms to augment the network architecture such that the result network reduces ReLU usages. Such works include DELPHI~\cite{srinivasan2019delphi} for which ReLUs are approximated with quadratic polynomial and a NAS algorithm is employed to automatically discover which ReLUs are replaced with the approximations. Another is CryptoNAS~\cite{ghodsi2020cryptonas} in which the authors developed a NAS over a fixed-depth, skip connections architectures to reduce ReLU count.

Two prominent state-of-the-art methods in this domain are DeepReDuce~\cite{jha2021deepreduce} and Selective Network Linearization (SNL)~\cite{cho2022selective}. DeepReDuce is composed of three sequential stages: ReLU "culling," ReLU "thinning," and ReLU "reshaping." However, these stages necessitate several manual design decisions and involve multiple hyper-parameters. In contrast, SNL~\cite{cho2022selective} introduces an individual parameter for each ReLU within the Neural Network. We refer to this parameter as the convex combination parameter. It governs whether a given ReLU is activated or replaced with an identity function. SNL employs a gradient-based approach to replace ReLUs with identity operations iteratively, solving for these parameters. During each iteration, the replacement process is coupled with the optimization of neural network parameters. Nevertheless, it's important to note that the final result network obtained through SNL's process requires re-training.

While the optimization-based method SNL displays promise by relaxing some of DeepReDuce's design constraints, our paper addresses the limitations it introduces. Unlike SNL, our approach handles networks that accommodate varying input image shapes, such as those encountered in Semantic Segmentation tasks. In contrast to SNL's stopping criterion, which often falls short of precisely attaining the desired non-linearity budget, our method consistently achieves an accurate budget. Additionally, SNL's iterative process necessitates network fine-tuning during each optimization iteration before arriving at a new Neural Network architecture. This considerably slows down their process compared to our inference and Knapsack-based pre-processing. Further elaboration on this topic can be found in the Supplementary Material.

In~\cite{helbitz2021reducing}, which is the work that has most influenced us, the authors used statistics from neighboring activations and shared DReLUs among them. However, they did not provide a satisfactory method for determining the neighborhood.
Finally, Circa~\cite{ghodsi2021circa} proposed the Stochastic ReLU layer and showed that we can use prior knowledge about the absolute size of activations to reduce the complexity of garbled circuits while maintaining a low error probability. They also demonstrated that neural networks can handle clipping of the least significant bits, which further reduces circuit complexity. While we share similar observations with Circa, our approximate ReLU layer differs in its application, as we ignore the shares' most significant bits instead of the negligible probability event of a shares summation overflow. This enables us to better control bandwidth with error probability, which is especially useful when reducing shares precision, as in ~\cite{wagh2020falcon}.

\paragraph{Network Pruning}
Network pruning is a technique applied to reduce model complexity by removing weights that contribute the least to model accuracy. Our method can be seen as a specialized pruning technique that targets DReLU operations instead of weights. Different approaches use various measures of importance to identify and eliminate the least important weights, such as magnitude-based pruning (e.g.,~\cite{han2015deep, han2015learning}), similarity and clustering methods (e.g., ~\cite{son2018clustering, roychowdhury2017reducing, dubey2018coreset}), and sensitivity methods (e.g.,~\cite{hassibi1992second}). Some methods, including ours, score filters based on the reconstruction error of a later layer (e.g.,~\cite{he2017channel, luo2017thinet, kamma2019reconstruction}). The pioneers of Knapsack based network pruning~\cite{aflalo2020knapsack} used the 0/1 Knapsack solution to eliminate filters in a neural network, whereby an item's weight was based on the number of FLOPs, and its value was based on the first-order Taylor approximation of the change to the loss. Similarly, in~\cite{shen2022structural} the authors optimized network latency using the Knapsack paradigm. The authors of~\cite{hubara2021accurate} applied Integer Programming to optimize post training quantization.

\section{Method}

Our approach is built upon the observation that ReLUs within a spatial neighborhood often exhibit similar non-linear activation responses. Leveraging this insight, we introduce Block-ReLU (bReLU), a non-linear activation function defined over a patch of activations. Substituting ReLUs with Block-ReLUs in the network reduces the number of non-linear operations, a desirable trait for Private Inference. However, this replacement introduces a certain degree of error. We aim at reaching a desired ReLU budget while minimizing the error incurred by our replacement.

To accomplish this, we develop a distortion measure that quantifies the degradation resulting from replacing ReLUs with Block-ReLUs of varying patch sizes. Assuming distortion additivity, we iterate ReLU channels in the Neural Netowrk. In each iteration we replace the \emph{current channel}'s ReLU with Block-ReLU. We evaluate the network distortion for different patch-sizes for that replacement. With these distortion values in hand, we formulate the optimal patch sizes as a variant of the Knapsack optimization problem. Solving this optimization yields the optimal patch sizes under our assumptions.

Finally, we fine-tune the neural network and apply the previously suggested bit-truncation~\cite{nishide2007multiparty, ghodsi2021circa} for yet another runtime and bandwidth consumption Secure Inference performance boost. For Secure Inference, we follow the SecureNN~\cite{wagh2019securenn} protocol, which is concisely summarized in the appendix.

\begin{figure}[t]
\begin{center}
   \includegraphics[width=0.6\linewidth]{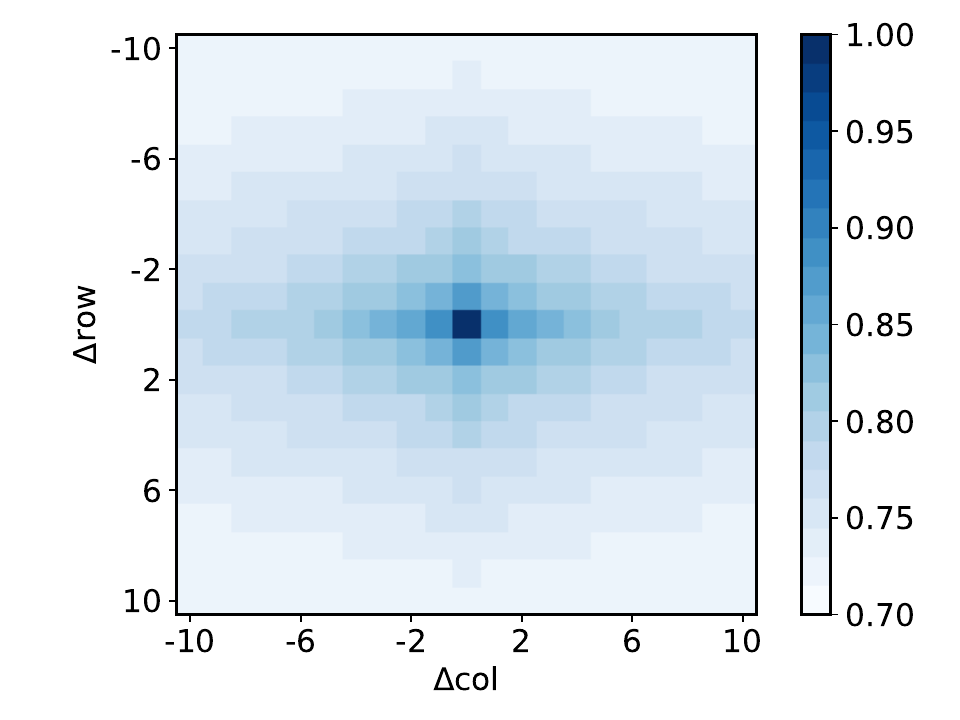}
\end{center}
   \caption{{\bf Activation Statistics for MobileNetV2:} The probability that two activation units in the same channel have the same sign, based on their spatial distance. The lowest probability is about $0.7=70\%$.}
\label{fig:pixel_correlation}
\end{figure}

\subsection{Approach}

Following SecureNN notations, we refer to the ReLU decision of an activation unit as its DReLU. Mainly:
\begin{equation*}
\text{DReLU}(x) = 
\begin{cases}
        1 & \text{if } x\geq 0\\
        0 & \text{otherwise}
    
\end{cases}
\end{equation*}
Then, ReLU is defined as:
\begin{equation}
\text{ReLU}(x) = x \cdot \text{DReLU}(x)
\end{equation}

\paragraph{Block-ReLU (bReLU)}
Our key insight is that neighboring activation units tend to have similar signs. To demonstrate this, we utilize a pre-trained Segmentation network trained on the ADE20K dataset with MobileNetV2 backbone. We evaluate the empirical probability of two activation units residing in the same channel having the same sign. Figure~\ref{fig:pixel_correlation} shows the probability map as a function of horizontal and vertical distances. Remarkably, even for $\text{lag}=10$, the probability remains at approximately $\sim 70\%$. To leverage this insight, we introduce the block ReLU (bReLU) layer.

Block ReLU utilizes the local spatial context of an activation unit to estimate its ReLU decision. Specifically, each activation channel is partitioned into rectangular patches of a specified size. The dimensions of the rectangle will be determined next using a Knapsack optimization solution.

Let $x$ be an activation unit, and $P(x)$ be its neighborhood induced by the rectangle partition. We define the patch ReLU decision (pDReLU) as the DReLU of the mean activation value of the neighborhood:

\begin{equation}
\text{pDReLU}(x) = \text{DReLU}(\frac{1}{|P(x)|}\Sigma_{a\in{P(x)}}{a})
\end{equation}

The ReLU decision of each pixel in the neighborhood is then replaced by the patch decision of that pixel.

\begin{equation}
\text{bReLU}(x) = x \cdot \text{pDReLU}(x)
\label{eq:bReLU}
\end{equation}

\begin{figure}
\begin{tabular}{c|c}
     Input Activation & Output Activation \\
     \includegraphics[width=0.45\linewidth]{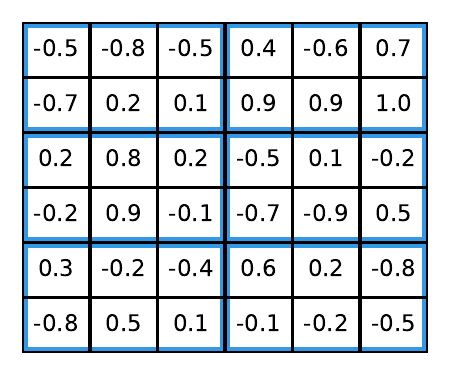} & 
     \includegraphics[width=0.45\linewidth]{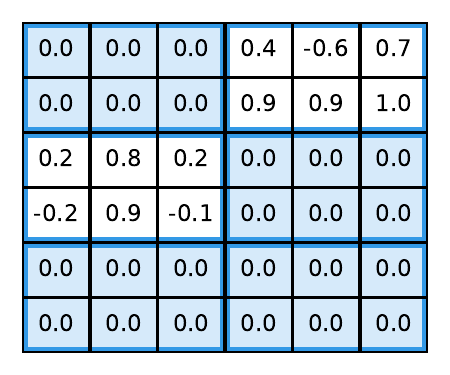}
\end{tabular}
\caption{\textbf{Block-ReLU (bReLU):} Left: $6 \times 6$ channel input, Right: bReLU output with a patch-size of $2 \times 3$. We convert $36$ DReLU operations to a mere $6$ operations, with the cost of $12$ sign flips.}

\label{fig:bReLU}
\end{figure}

As a result, we perform a single DReLU operation in the $P(x)$ patch, instead of $|P(x)|$ individual DReLU operations. This is our major contribution in terms of cutting the number of non-linear operations. The extent in which we save non-linear activations is determined by the patch area (or neighborhood size) $|P(x)|$. Figure~\ref{fig:bReLU} illustrates a bReLU with a patch size of $2 \times 3$ operating on a $6 \times 6 $ activation channel.

\paragraph{Knapsack Optimization}
Our goal is  to adhere to some DReLU budget $\mathcal{B}$, with minimal degradation to the network performance. To achieve this, we solve for the Block-ReLU patch size.  We draw inspiration from the Rate-Distortion~\cite{berger2003rate} theory. We devise a distortion measure which serves as a proxy for the network performance and the rate is addressed as the ratio between the number of bReLUs to the number of ReLUs: $\frac{\#\text{bReLUs}}{\# \text{ReLUs}}$. 

Different layers in the neural correspond to different feature hierarchies and different channels within the same layer correspond to different features. Therefore, determining a neighbourhood size or patch-size for the bReLU operation is \emph{channel} dependent. We enumerate the channels across the network and denote $C_i$ the $i^{th}$ channel in a neural network. For example, in a network consisting of two layers, with $32$ channels in the first layer and $64$ in the second layer, $C_{40}$ would refer to the $8^{th}$ channel in the second layer.

Next, we define $P_{i}$ as the list of patch-sizes available for selection by $C_{i}$, and $P_{ij}$ as the $j$-th item within this list. $\mathcal{D}(i, j)$ is then defined as the distortion caused to the network as a result of replacing the ReLUs of $C_{i}$ with a bReLU of patch size  $P_{ij}$.

The parameter we optimize is the patch-size $P_{ij}$. Evaluating the network's performance for each patch-size configuration is exponentially complex with the number of patches (proportional to $L^{k}$, where $L$ is the number of channels in the network and $k$ is the number of patch-size options).

To address this challenge, we introduce two relaxations for this task. First, we devise a distortion measure to serve as a proxy for the network performance degradation. We choose to measure this distortion by quantifying the change in the network activation for the same inputs. This makes distortion a general purpose solution for both classification and Semantic Segmentation. Clearly, future work can discuss other types of distortion methods. 

Formally, let $F$ be a neural network, and $F^{i, j}$ denote the network obtained by replacing the ReLUs of $C_{i}$ of $F$ with bReLU of a patch size $P_{ij}$.  Furthermore, let $F(X)$ and $F^{i, j}(X)$ be the last activation layers of $F$ and $F^{i, j}$ operating on an image $X$. Then the distortion $\mathcal{D}(i, j)$ is defined as follows:
\begin{equation}
\mathcal{D}(i, j) = \mathop{\mathbb{E}_{X}}[\lVert F^{i, j}(X) - F(X) \rVert^{2}]
\label{eq:distortion}
\end{equation}
The distortion is calculated as the average $L_{2}$ distance between the last activation layers of the original pre-trained network and a similar network, which has the same weights but replaces the ReLU in $C_{i}$ with a bReLU having a patch-size of $P_{ij}$. This measure is evaluated across a subset of train images. 

The second relaxation that we make is to approximate the total distortion resulting from replacing ReLUs with bReLUs for a number of channels as the additive distortion. This allows us to cast the optimal patch-shape as a Knapsack optimization problem. The cost for optimization problem is determined by the number of DReLUs remaining in $C_{i}$ of $F^{i, j}$, which we denote as $\mathcal{W}(i, j)$. It is important to note that $\mathcal{W}(i, j)$ is influenced by both the patch-size and the activation channel size. Additionally, we define $m$ as the number of channels in $F$, and $S_{i}$ as the set of indices ${1, ..., |P_{i}|}$. With these notations, our optimization problem boils down to the Multiple-Choice Knapsack Problem~\cite{kellerer2004multiple}:

\begin{equation}
\begin{aligned}
\min_{\varphi_{ij}} \quad & \sum_{i=1}^{m}{\sum_{j\in{S_{i}}}\mathcal{D}(i, j)\cdot{\varphi_{ij}}}\\
\textrm{subject to} \quad & \sum_{i=1}^{m}{\sum_{j\in{S_{i}}}\mathcal{W}(i, j)\cdot{\varphi_{ij}}} \leq{\mathcal{B}} ,\\
  &\sum_{j\in{S_{i}}}{\varphi_{ij}} =1, \quad i=1,...,m,    \\
  & \varphi_{ij}\in\{0,1\}, \quad i=1,...,m, \quad j\in{S_{i}}  \\
\end{aligned}
\label{eq:optimization}
\end{equation}

The indicator variable $\varphi_{ij} = 1$ indicates that the $P_{ij}$ patch size has been selected for $C_{i}$. The second constraint ensures that only a single patch size can be selected for this channel, thereby preventing the selection of multiple patch sizes.

We use Dynamic Programming to solve this optimization problem. We iteratively populate a table $DP$ consisting of $m$ rows and $\mathcal{B}$ columns, using the following recursive formula:

\begin{equation}
DP[i,j] = \min_{1\leq{l}\leq{S_{i}}} \{DP[i-1, j-\mathcal{W}(i, l)] + \mathcal{D}(i, l)\}
\end{equation}

As this table is column-independent, we can parallelize the process of finding the solution $DP[m, \mathcal{B}]$. To fully leverage this property, we implement a GPU version of the algorithm. The code for the parallelized solver will be made publicly available upon acceptance.

\paragraph{Analysis} Our optimization cost in Equation~\ref{eq:optimization} assumes that the total distortion replacing ReLUs with bReLUs across multiple channels is additive. This implies that the distortion caused by the ReLU-bReLU replacement in two channels equals to the sum of distortions caused by the replacement in each channel individually. Notably, this assumption can be relaxed, with a sufficient condition being a monotonically non-decreasing relationship between the real distortion and the additive distortion. Further exploration of this relationship is detailed in the supplementary material.

\paragraph{Homogeneous Channels}
Some channels tend to have uniform signs across the entire channel, where all activations are either positive or negative. To leverage this observation, we extended the $P_{i}$ lists, which hold the different options for patch-sizes, by adding an additional special item: the identity channel. This identity channel has a weight of $\mathcal{W}=0$, indicating that no DReLUs are employed in this case.

\subsection{Secure Inference Considerations}
\paragraph{Approximate DReLU} In protocols that utilize additive sharing, the cost of a comparison operation is typically proportional to the number of bits used to represent the shares. This relationship is evident in the depth of a Yao's Garbled Circuit, and in SecureNN-based protocols. Previous observations~\cite{nishide2007multiparty, ghodsi2021circa} indicate that by allowing for a small probability of DReLU error, the complexity of comparison operations can be substantially reduced. We analyze the probability of a DReLU error as a function of the number of most and least significant bits ignored. We use ResNet50 with bReLU layers trained on the ImageNet dataset by and examine millions of activation values. Based on this analysis, we choose to disregard 43 of the most significant bits and 5 of the least significant bits, resulting in a DReLU error probability of about $5 \times 10^{-4}$.

\paragraph{MaxPool and ReLU6}
MaxPool and ReLU6 are computationally costly layers for secure inference. We suggest to substitute ResNet's MaxPool layer with an AveragePool layer and MobileNet's ReLU6 layers with ReLU layers. If required, we also adjust the models through fine-tuning. Exact fine-tuning details are summarized in the supplemental.

\subsection{Security Analysis}
Our method offers the same level of security as the underlying cryptographic protocol, SecureNN in our case. Accordingly, we protect both the client image and server model weights under the relevant thread model. However, we do make the patch sizes public, which typically consists of approximately 40K discrete parameters that we infer based on the training data statistics. It is important to note that any hyperparameters or architectural structures that are revealed may result in some degree of training data information leakage. Therefore, researchers should be aware of this gray area and determine what level of model disclosure is acceptable. Other than image size, the user's image information is entirely protected.
\section{Experiments}

\paragraph{Implementation Details}
To evaluate the efficacy of our method, we use the standard Classification benchmark classifying CIFAR-100~\cite{krizhevsky2010cifar} with the ResNet18 backbone. We also evaluate the classification of ImageNet~\cite{deng2009imagenet}, using ResNet50~\cite{he2016deep} backbone and perform an evaluation on two Semantic Segmentation datasets: ADE20K~\cite{zhou2017scene} using DeepLabV3~\cite{chen2017rethinking} with MobileNetV2~\cite{sandler2018mobilenetv2} backbone and Pascal VOC 2012~\cite{everingham2010pascal} using DeepLabV3 and ResNet50 backbone. Pre-trained networks are obtained from the OpenMMLab model zoo, other than the missing CIFAR100 models which were trained from scratch.

\begin{figure*}[t]
\begin{tabular}{cc || cc}
        \multicolumn{2}{c||}{Secure Inference Performance} & \multicolumn{2}{c}{Components Contribution}\\

    \includegraphics[width=0.22\linewidth]{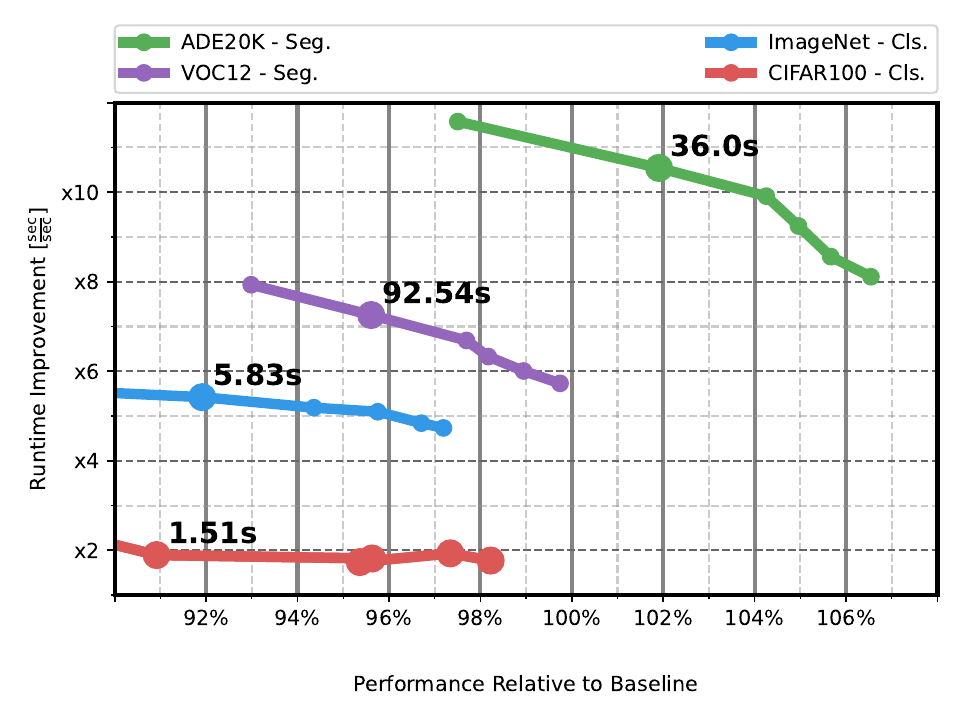} & 
    \includegraphics[width=0.22\linewidth]{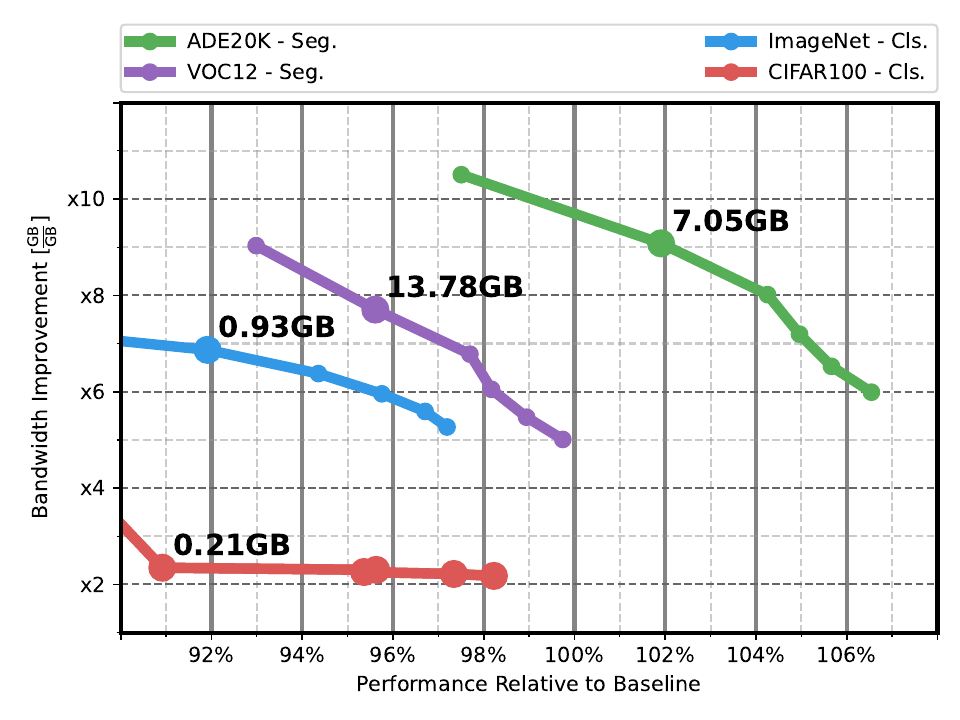} &
    \includegraphics[width=0.22\linewidth]{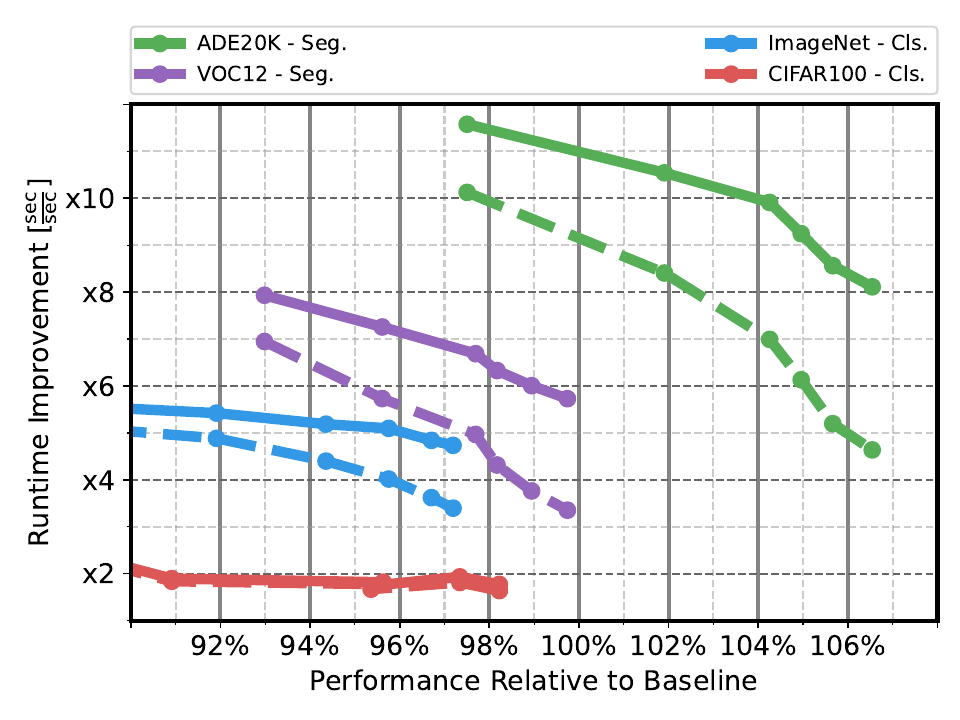} &
    \includegraphics[width=0.22\linewidth]{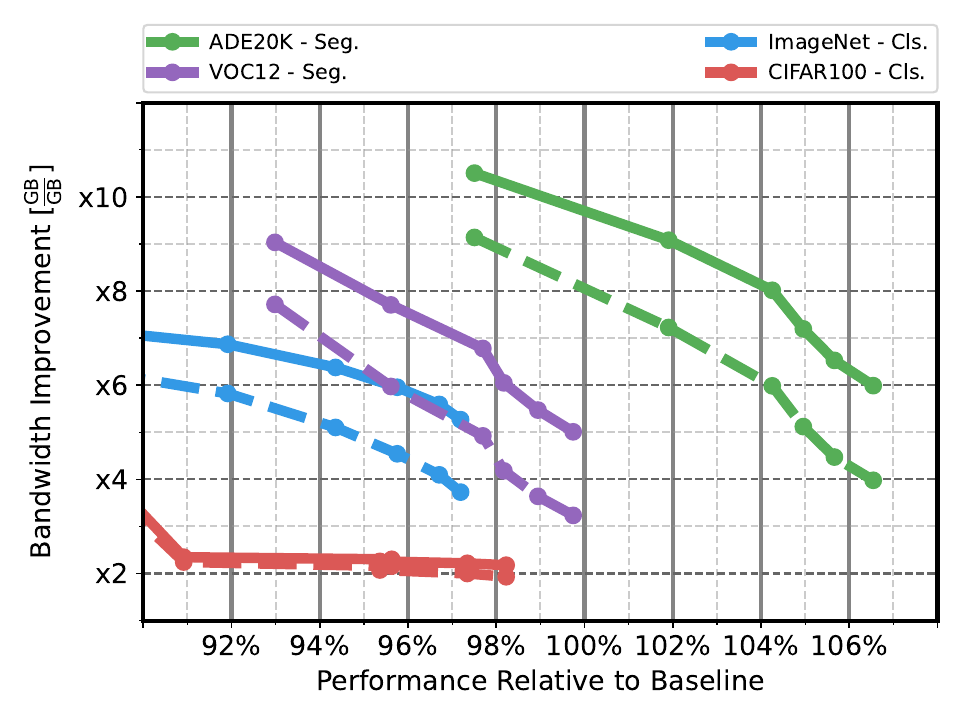} \\
    (a) Runtime & (b) Bandwidth & (c) Runtime & (d) Bandwidth \\

\end{tabular}

   \caption{{\bf Secure Inference:} Runtime (a) and Bandwidth consumption (b) improvement are depicted vs the task performance operating point. The baseline model comprises the pre-trained version with all ReLUs and full ReLU resolution. Our evaluation employs the SecureNN protocol, conducted on a cloud environment. Specific runtime and bandwidth consumption values are explicitly provided for operting points denoted with larger circles. Figures (c) and (d) show the contribution of each component to the Secure Inference runtime and bandwidth consumption metrics. Solid lines are evaluations with Approximate DReLU, while the dashed lines indicate the network with bReLUs evaluated without approximate DReLUs.}

\label{fig:runtime}
\end{figure*}

We evaluate distortion through forward passes in the Neural Network. The distortions are evaluated for the different patch-sizes and channels. The optimal patch-sizes were determined using our CUDA-based Multiple-Choice-Knapsack solver. ReLU layers were replaced with bReLU layers, parameterized by the Knapsack-optimal patch-sizes, and the models were fine-tuned. Details of the Knapsack search parameters (e.g. patch-sizes lists), fine-tuning hyper parameters  and computing infrastructure are provided in the supplementary material. 

\paragraph{Evaluation}
We perform two kinds of evaluations: (1) we measure the task performance of our secure network and compare it to DeepReDuce and the SOTA SNL. We also present the baseline non-secure task performance of the Neural Network on the original task (Classification or Semantic Segmentation), (2) we evaluate runtime and bandwidth consumption of our network under SecureNN, a specific Secure Inference protocol. 

Figure~\ref{fig:teaser} shows the tradeoff between ReLU savings and task performance. We evlauate our method against DeepReDuce and SNL for the classification of CIFAR-100 and against SNL for ImageNet. We use black dashed line to indicate the non-secure baseline task performance. For almost all classification benchmarks, we outperform competitors. It is important to highlight that SNL in it's current form, cannot be used on variable size images and is therefore not evaluated for the Semantic Segmentation benchmark.

The second type of evaluation we conduct is a Secure Inference protocol test which measures the communication bandwidth and latency (runtime) of a secure protocol using our suggested secure network. For this, we set up 3 Amazon EC2 c4.8xlarge Ubuntu-running instances in the same region (eu-west-1). We developed our own Numba~\cite{lam2015numba} based Python implementation of the SecureNN protocol. Further secure inference implementation details are in the supplementary material. Runtime performance is evaluated for $3$ images for Semantic Segmentation and $10$ images for classification tasks. Baseline models are downloaded from the OpenMMLab model zoo and include neither bReLU nor approximate ReLU components. 

To assess the effectiveness of our network in terms of Secure Inference, we evaluate how it reduces runtime and bandwidth consumption compared to a baseline model containing all ReLUs and utilizing full ReLU resolution. While Figure~\ref{fig:teaser} shows the performance of the network on the task vs. the DReLU  budget, Figures~\ref{fig:runtime}(a)+(b) shows the runtime and communication bandwidth savings of the entire protocol compared to the non-secure version. Figures~\ref{fig:runtime}(a), (b) illustrate the runtime and bandwidth savings achieved by our network in comparison to the baseline model. Each point on the graph corresponds to a distinct operating point in terms of task performance, with mIoU for Semantic Segmentation and Accuracy for Classification. Lower task performance corresponds to a reduced DReLU budget. Higher improvement correspond to better runtime or bandwidth communication consumption performance over the baseline non secure model. The graph reflects the tradeoff between task performance and runtime and bandwidth consumption. For completeness, we compare secure and non secure performance in Table~\ref{tab:secure_non_secure_comparison}, under a fixed DReLU budget of $6\%$.

\begin{table}
\begin{adjustbox}{width=0.47\textwidth}

\begin{tabular}{|l|c|c|c|c|}
\hline
~ & \multicolumn{2}{c|}{Classification} & \multicolumn{2}{c|}{Segmentation} \\

~ & ImageNet & CIFAR-100 & ADE20K & VOC12  \\
~ & ResNet50 & ResNet18 & MobileNetV2 & ResNet50  \\
\hline\hline
Non-secure & 70.36 & 70.90 & 34.73 & 74.17 \\
\hline
Ours (Secure) & 70.27 & 71.16 & 34.64 & 74.21 \\

\hline
\end{tabular}

\end{adjustbox}
\caption{{\bf Secure vs. Non-secure Task Performance} With a DReLU budget of 6\%, the task performance, measured by mIoU for Semantic Segmentation and Accuracy for Classification, remains largely unchanged switching from Non-Secure to our (Secure) network.}
\label{tab:secure_non_secure_comparison}
\end{table}

\paragraph{Secure Inference Analysis} Table~\ref{tab:Communication} displays the cost of communication for the various layers in the SecureNN protocol, with the CrypTFlow convolution optimization being utilized. 
We define $h$ as the activation dimension. $i$ and $o$ are the number of activation input and output channels. $f$ is the convolution kernel size. $\ell$ is the number of bits used to represent activation values. $\ell^{*}$ is the number of bits used in approx. DReLU. $P$ is a list of patch-sizes such that $|P| = o$. We set $r=6$log$p\ell + 14\ell$ (3,968 in our case), $r^{*}=6$log$p\ell^{*} + 14\ell$ (1,664 in our case) and $q=\frac{1}{o}\sum_{i=1}^{o}\frac{1}{P{i}}$, the ratio of DReLUs left. The communication cost of ReLU and bReLU is defined as the communication that remains after DReLU and pDReLU have been applied. The communication cost of some typical values ($i=128, o=256, f=3, h=64, \ell=64, \ell^{*}=16,  q = 0.1$) is shown. The complexity reduction caused by using identity channels is not shown. Finally, theoretically, in approximate DReLU layer, we can further reduce communication by decreasing the $\mathbb{Z}_{p}$ field size, as the only requirement is that $p$ is a prime such that: $p> 2+\ell^{*}$.

The usage of bReLU layer does not decrease the number of communication rounds required in comparison to a standard ReLU layer. This holds true irrespective of whether or not approximate DReLU is utilized. As a result, the bReLU layer incurs a cost of 10 communication rounds. We refer the readers to~\cite{wagh2019securenn} for more details.

\begin{table}
\begin{center}
\begin{tabular}{|l|c|c|c|}
\hline
~ & Layer & Communication & Typical   \\
~ & ~ &  ~ & Values  \\
\hline\hline
1 & Conv2d$_{h,i,f,o}$ & $h^{2}(2i+o)\ell$ & $21$MB \\
~ & ~ & $+2f^{2}oi\ell$ & ~ \\
\hline
2 & DReLU$_{h,o}$ & $h^{2}ro   $ & $520$MB \\
3 & App. DReLU$_{h,o}$ & $h^{2}r^{*}o   $  & $218$MB \\
\hline
4 & pDReLU$_{h,o,P}$ & $h^{2} r qo$  & $52$MB \\
5 & App. pDReLU$_{h,o,P}$ & $h^{2}r^{*}qo  $  & $22$MB \\
\hline
6 & ReLU$_{h,o}$ & $5oh^{2}\ell  $   & $42$MB \\
7 & bReLU$_{h,o, P}$ & $(3+2q)oh^{2}\ell  $   & $27$MB \\
\hline\hline
8 & Conv2d+ReLU & (1) + (2) + (6) & $583$MB \\
9 & Conv2d+bReLU & (1) + (5) + (7) & $70$MB \\
\hline
\end{tabular}
\end{center}
\caption{{\bf Communication compexity:} Communication compexity of Conv2D and ReLU layers. Our approximation (line (9)) requires almost an order of magnitude less communication bandwidth, compared to the baseline approach (line (8)). See discussion and details in the text. } 

\label{tab:Communication}
\end{table}

\subsection{Ablation Study}\label{sec:ablation}

\paragraph{Alternative Patch Sizes Sets}
We investigate the effect on performance of using a different set of patch-sizes instead of the Knapsack optimal patch-sizes.
Table~\ref{tab:PatchSets} presents a comparison of three different patch size sets: (1) A set of naive $4 \times 4$ constant patch sizes, (2) Channel-shuffled Knapsack patch sizes, which are similar to Knapsack-optimized patch sizes but are shuffled among channels within the same layer, and (3) Knapsack-optimized patch sizes using the same DReLU budget as the constant version. The purpose of the third set is to differentiate between the respective contributions of the coarse-grained budget allocation across layers and the fine-grained budget allocation across channels within the same layer.
We note that shuffling preserves the patch size layer distribution, and thus only partially isolate the contribution of this fine grained allocation.

\begin{table}
\begin{center}
\begin{tabular}{|l|c|c|c|c|}
\hline
~ & ImgNt & CIFR100 & AD20K & VOC12  \\
~ & ResN50 & ResN18 & MbNV2 & ResN50  \\
\hline\hline
$4\times 4$ & 60.59 & 59.07 & 31.06 & 67.64 \\
\hline
KS-Shuffled & 69.55 & 70.67 & 33.20 & 72.95 \\
\hline
KS-Optimal & {\bf 71.04} & {\bf 71.28} & {\bf 35.08} & {\bf 74.91} \\
\hline
\end{tabular}
\end{center}
\caption{{\bf Alternative Patch Size Sets} The effect of using different sets of patch sizes at a given budget of DReLUs on task performance. \emph{KS-shuffled} refers to shuffling the set of Knapsack optimal patch sizes within layers. \emph{KS-Optimal} refers to using the set of Knapsack-optimal patch sizes.}
\label{tab:PatchSets}
\end{table}

\paragraph{Approximate DReLU} 
Figures~\ref{fig:runtime}(c) + (d) breaks down the contribution of solely bReLUs (dashed lines) and adding the Approximate DReLUs on top of them. Largely, both bReLUs and Approximate DReLUs contribute to the metrics with bReLUs being more significant.

\section{Conclusions}
We propose a technique to decrease DReLU counts in Neural Networks, aiming to enhance Secure Inference runtime and bandwidth usage. Our key insight is that we can replace ReLU activations with Block-ReLU(bReLU) activations which operate on a patch. Utilizing our devised distortion measure, we formulate the problem of finding the optimal bReLU patch-size as a Knapsack optimization problem. Our formulation allows us to precompute distortions \emph{once} and then reuse them to find the optimal bReLU patch sizes for different DReLU budgets. This flexibility allows us to explore the tradeoff between task performance and DReLU utilization.

We showcase our approach's competitiveness on the standard CIFAR-100 classification benchmark using ResNet18 and surpass the current state-of-the-art method, SNL, in ImageNet classification with ResNet50. We are the \emph{first} to demonstrate Secure Inference on the Semantic Segmentation task. We evaluate runtime and bandwidth consumption for a secure inference setting implemented within the 3-party SecureNN protocol.

\bibliography{aaai24}

\end{document}